\documentclass[final]{cvpr}

\usepackage{times}
\usepackage{epsfig}
\usepackage{graphicx}
\usepackage{amsmath}
\usepackage{amssymb}
\usepackage[toc,page,title]{appendix}
\BeforeBeginEnvironment{appendices}{\clearpage}
\usepackage{textcomp}
\usepackage{xcolor}
\usepackage{gensymb}
\usepackage{multirow}
\usepackage[linesnumbered,ruled]{algorithm2e}
\usepackage{balance}
\SetAlFnt{\small}
\usepackage[font=small,skip=0pt]{caption}
\newcommand{\norm}[1]{\left\lVert#1\right\rVert_{1}}
\newcommand{\normsecond}[1]{\left\lVert#1\right\rVert_{2}}
\usepackage[pagebackref=true,breaklinks=true,colorlinks,bookmarks=false]{hyperref}

\def\cvprPaperID{261} 
\def\confYear{CVPR 2021}

\begin{document}
\title{Joint Generative and Contrastive Learning for Unsupervised Person Re-identification \vspace{-10pt}}


\author{Hao Chen\textsuperscript{1,2,3}\thanks{Equal contribution.
} \hskip 1em 
Yaohui Wang\textsuperscript{1,2}\footnotemark[1] \hskip 1em
Benoit Lagadec\textsuperscript{3} \hskip 1em 
Antitza Dantcheva\textsuperscript{1,2} \hskip 1em 
Francois Bremond\textsuperscript{1,2}\\
\textsuperscript{1}Inria \hskip 1em
\textsuperscript{2}Université Côte d'Azur \hskip 1em
\textsuperscript{3}European Systems Integration\\
{\tt\footnotesize \{hao.chen, yaohui.wang, antitza.dantcheva, francois.bremond\}@inria.fr} \hskip 1em 
{\tt\footnotesize benoit.lagadec@esifrance.net} \\
\vspace{-25pt}
}

\maketitle
\begin{abstract}
Recent self-supervised contrastive learning provides an effective approach for unsupervised person re-identification (ReID) by learning invariance from different views (transformed versions) of an input. 
In this paper, we incorporate a Generative Adversarial Network (GAN) and a contrastive learning module into one joint training framework. While the GAN provides online data augmentation for contrastive learning, the contrastive module learns view-invariant features for generation.
In this context, we propose a mesh-based view generator. Specifically, mesh projections serve as references towards generating novel views of a person. In addition, we propose a view-invariant loss to facilitate contrastive learning between original and generated views. Deviating from previous GAN-based unsupervised ReID methods involving domain adaptation, we do not rely on a labeled source dataset, which makes our method more flexible. Extensive experimental results show that our method significantly outperforms state-of-the-art methods under both, fully unsupervised and unsupervised domain adaptive settings on several large scale ReID datsets. Source code and models are available under \url{https://github.com/chenhao2345/GCL}. \vspace{-15pt}
\end{abstract}

\section{Introduction}

A person re-identification (ReID) system is targeted at identifying subjects across different camera views. In particular, given an image containing a person of interest (as query) and a large set of images (gallery set), a 
ReID system ranks gallery-images based on visual similarity with the query.
Towards this, ReID systems are streamlined to bring to the fore discriminative representations, which allow for robust comparison of query and gallery images. In this context, \textit{supervised} ReID methods \cite{Chen_2020_WACV,sun2018beyond} learn representations guided by human-annotated labels, which is 
time-consuming and cumbersome. Towards omitting such human annotation, researchers increasingly place emphasis on \textit{unsupervised} person ReID algorithms \cite{Wang_2020_CVPR,li2020joint,Lin2019ABC}, which learn directly from unlabeled images and thus allow for scalability in real world deployments. 

\begin{figure}
\centering
  \includegraphics[width=0.9\linewidth]{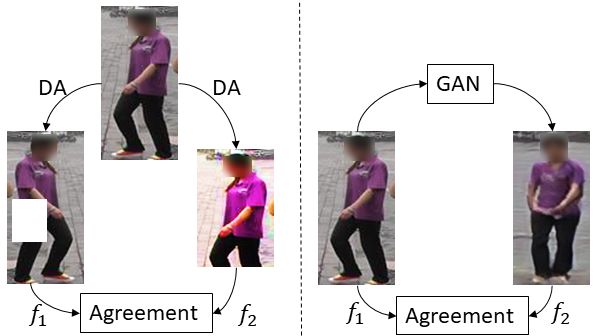}
  \caption{\textbf{Left}: Traditional self-supervised contrastive learning maximizes  agreement between representations ($f_1$ and $f_2$) of augmented views from Data Augmentation (DA). \textbf{Right}: Joint generative and contrastive learning maximizes agreement between original and generated views.}
\label{fig:traditional vs gan}
\end{figure}

Recently, self-supervised contrastive methods \cite{He_2020_CVPR,chen2020simple} have provided an effective retrieval-based approach for unsupervised representation learning. Given an image, such methods maximize agreement between two augmented views of one instance (see Fig.~\ref{fig:traditional vs gan}). \textit{Views} refer to transformed versions of the same input. 
As shown in very recent works \cite{chen2020simple,chen2020improved}, data augmentation enables a network to explore view-invariant features by providing augmented views of a person, which are instrumental in building robust representations. Such and similar methods considered traditional data augmentation techniques, \eg, `random flipping', `cropping', and `color jittering'. Generative Adversarial Networks (GANs) \cite{goodfellow2014generative} constitute a novel approach for data augmentation. 
As opposed to traditional data augmentation, GANs are able to modify id-unrelated features substantially, while preserving id-related features, which is highly beneficial in contrastive ReID.

Previous GAN-based methods \cite{bak2018domain,deng2018image,Zou2020JointDA,li2019cross,wei2018person,Zhong_2018_ECCV} considered unsupervised ReID as an unsupervised domain adaptation (UDA) problem. Under the UDA setting, researchers used both, a labeled source dataset, as well as an unlabeled target dataset to gradually adjust a model from a source domain into a target domain. GANs can be used in cross-domain style transfer, where labeled source domain images are generated in the style of a target domain. However, the UDA setting necessitates a large-scale labeled source dataset. Scale and quality of the source dataset strongly affect the performance of UDA methods. Recent research has considered fully unsupervised ReID \cite{Wang_2020_CVPR,li2020joint}, where under the fully unsupervised setting, a model directly learns from unlabeled images without any identity labels. Self-supervised contrastive methods \cite{He_2020_CVPR,chen2020simple} belong to this category. In this work, we use a GAN as a novel view generator for contrastive learning, which does not require a labeled source dataset.

Here, we aim at enhancing view diversity for contrastive learning via generation under the fully unsupervised setting. Towards this, we introduce a mesh-based novel view generator. We explore the possibility of disentangling a person image into identity features (color distribution and body shape) and structure features (pose and view-point) under the fully unsupervised ReID setting. We estimate 3D meshes from unlabeled training images, then rotate these 3D meshes to simulate new structures. Compared to skeleton-guided pose transfer \cite{ge2018fd,li2019cross}, which neglects body shape, mesh recovery \cite{kanazawaHMR18} jointly estimates pose and body shape. Estimated meshes preserve body shape during the training, which facilitates the generation and provides more visual clues for fine-grained ReID. Novel views can be generated by combining identity features with new structures. 

Once we obtain the novel views, we design a pseudo label based contrastive learning module. With the help of our proposed view-invariant loss, we maximize representation similarity between original and generated views of a same person, whereas representation similarity of other persons is minimized. 

Our proposed method incorporates generative and contrastive modules into one framework, which are trained jointly. Both modules share the same identity feature encoder. The generative module disentangles identity and structure features, then generates diversified novel views. The novel views are then used in the contrastive module to improve the capacity of the shared identity feature encoder, which in turn improves the generation quality. Both modules work in a mutual promotion way, which significantly enhances the performance of the shared identity feature encoder in unsupervised ReID. Moreover, our method is compatible with both UDA and fully unsupervised settings. With a labeled source dataset, we obtain better performance by alleviating the pseudo label noise. 

Our contributions can be summarized as follows.
\begin{enumerate}
  \item We propose a joint generative and contrastive learning framework for unsupervised person ReID. Generative and contrastive modules mutually promote each other's performance. 
  \item In the generative module, we introduce a 3D mesh based novel view generator, which is more effective in body shape preservation than skeleton-guided generators.
  \item In the contrastive module, a view-invariant loss is proposed to reduce intra-class variation between original and generated images, which is beneficial in building view-invariant representations under a fully unsupervised ReID setting.
  \item We overcome the limitation of previous GAN-based unsupervised ReID methods that strongly rely on a labeled source dataset. Our method significantly surpasses the performance of state-of-the-art methods under both, fully unsupervised, as well as UDA settings. 
\end{enumerate}

\section{Related Work}
\paragraph{Unsupervised representation learning.} 
Recent contrastive instance discrimination methods \cite{Wu2018UnsupervisedFL, He_2020_CVPR, chen2020simple} have witnessed a significant progress in unsupervised representation learning. The basic idea of instance discrimination has to do with the assumption that each image is a single class. Contrastive predictive coding (CPC) \cite{oord2018representation} included an InfoNCE loss to measure the ability of a model to classify positive representation amongst a set of unrelated negative samples, which has been commonly used in following works on contrastive learning. Recent contrastive methods treated unsupervised representation learning as a retrieval task. Representations can be learnt by matching augmented views of a same instance from a memory bank \cite{Wu2018UnsupervisedFL,He_2020_CVPR} or a large mini-batch \cite{chen2020simple}.
MoCoV2 \cite{chen2020improved} constitutes the improved version of the MoCo \cite{He_2020_CVPR} method, incorporating larger data augmentation. We note that data augmentation is pertinent in allowing a model to learn robust representations in contrastive learning. However, only traditional data augmentation was used in aforementioned methods. 

\paragraph{Data augmentation.} 
MoCoV2 \cite{chen2020improved} used `random crop', `random color jittering', `random horizontal flip', `random grayscale' and `gaussian blur'. However, `random color jittering' and `grayscale' were not suitable for fine-grained person ReID, because such methods for data augmentation tend to change the color distribution of original images. In addition, `Random Erasing' \cite{Zhong2020RandomED} has been a commonly used technique in person ReID, which randomly erases a small patch from an original image. Cross-domain Mixup \cite{luo2020generalizing} interpolated source and target domain images, which alleviated the domain gap in UDA ReID. 
Recently, Generative Adversarial Networks (GANs)~\cite{goodfellow2014generative} have shown great success in image ~\cite{Karras2019stylegan2, karras2019style, brock2018large} and video synthesis~\cite{tulyakov2017mocogan, wang2020g3an, chan2019everybody,wang2021inmodegan,9093492}. GAN-based methods can serve as a method for evolved data augmentation by conditionally modifying id-unrelated features (style and structure) for supervised ReID.
CamStyle \cite{Zhong2018CameraSA} used the CycleGAN-architecture \cite{CycleGAN2017} in order to transfer images from one camera into the style of another camera.
FD-GAN \cite{ge2018fd} was targeted to generate images in a pre-defined pose, so that images could be compared in the same pose. IS-GAN \cite{NIPS2019_8771} was streamlined to disentangle id-related and id-unrelated features by switching both local and global level identity features. DG-Net \cite{zheng2019joint} recolored grayscale images with a color distribution of other images, targeting to disentangle identity features. 
Deviating from such supervised GAN-based methods, our method generates novel views by rotating 3D meshes in an \textit{unsupervised} manner. 

\paragraph{Unsupervised person ReID.} 
Recent unsupervised person ReID methods were predominantly based on UDA. Among UDA-based methods, several works \cite{Wang2018TransferableJA, Lin2018MultitaskMF} used semantic attributes to facilitate domain adaptation. Other works \cite{wu2019progressive,fu2019self,Chen_2021_WACV,yang2020asymmetric,ge2020mutual} assigned pseudo labels to unlabeled images and proceeded to learn representations with pseudo labels. Transferring source dataset images into the style of a target dataset represents another line of research. SPGAN \cite{deng2018image} and PTGAN \cite{wei2018person} used CycleGAN \cite{CycleGAN2017} as domain style transfer-backbone.
HHL \cite{Zhong_2018_ECCV} aims at transferring cross-dataset camera styles. ECN \cite{zhong2019invariance,zhong2020learning} exploited invariance from camera style transferred images for UDA ReID. CR-GAN \cite{chen2019instance} employed parsing-based masks to remove noisy backgrounds. PDA \cite{li2019cross} included skeleton estimation to generate person images with different poses and cross-domain styles. DG-Net++ \cite{Zou2020JointDA} jointly disentangled id-related/id-unrelated features and transferred domain styles. While the latter is related to our our method, we aim at training jointly a GAN-based online data augmentation, as well as a contrastive discrimination, which renders the labeled source dataset unnecessary, rather than transferring style.

Fully unsupervised methods do not require any identity labels. BUC \cite{Lin2019ABC} represented each image as a single class and gradually merged classes. In addition, TSSL \cite{wu2020tracklet} considered each tracklet as a single class to facilitate cluster merging. SoftSim \cite{Lin2020UnsupervisedPR} utilized similarity-based soft labels to alleviate label noise. MMCL \cite{Wang_2020_CVPR} assigned multiple binary labels and trained a model in a multi-label classification way. JVTC and JVTC+ \cite{li2020joint} added temporal information to refine visual similarity based pseudo labels. We note that all aforementioned fully unsupervised methods learn from pseudo labels. We show in this work that disentangling view-invariant identity features is possible in fully unsupervised ReID, which can be an add-on to boost the performance of previous pseudo label based methods. 


\begin{figure*}
\centering
   \includegraphics[width=0.9\linewidth]{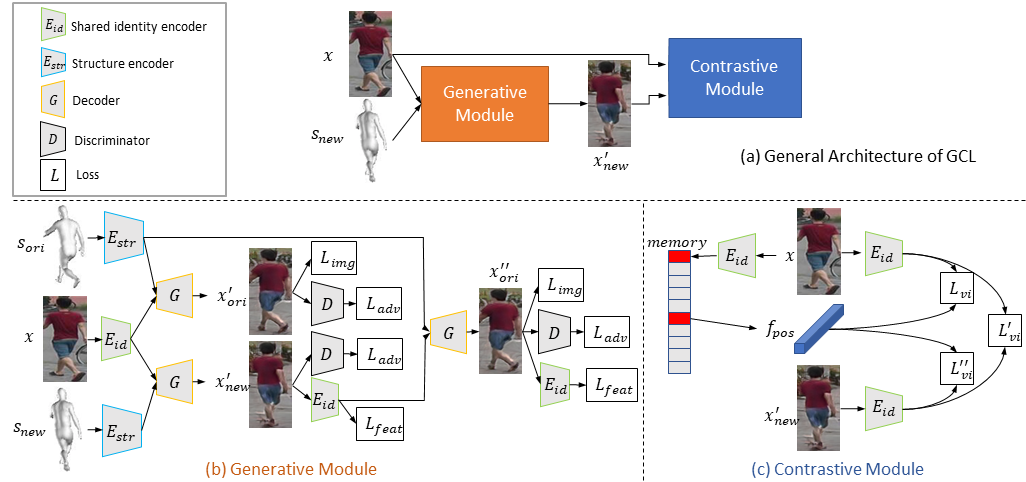}
   \caption{\textbf{(a) General architecture of GCL}: Generative and contrastive modules are coupled by the shared identity encoder $E_{id}$. \textbf{(b) Generative module}: The decoder $G$ combines the identity features encoded by $E_{id}$ and structure features $E_{str}$ to generate a novel view $x_{new}^{\prime}$ with a cycle consistency. \textbf{(c) Contrastive module}: View-invariance is enhanced by maximizing the agreement between original $E_{id}(x)$, synthesized $E_{id}(x_{new}^{\prime})$ and memory $f_{pos}$ representations.}
\label{fig:general structure}
\vspace{-5pt}
\end{figure*}

\section{Proposed Method}
We refer to our proposed method as joint\textit{ Generative and Contrastive Learning } as GCL. The general architecture of GCL comprises of two modules, namely a View Generator, as well as a View Contrast Module, see Fig. \ref{fig:general structure}. Firstly, the View Generator uses cycle-consistency on both, image and feature reconstructions in order to disentangle identity and structure features. It combines identity features and mesh-guided structure features to generate one person in new view-points. Then, original and generated views are exploited as positive pairs in the View Contrast Module, which enables our network to learn view-invariant identity features. We proceed to elaborate on both modules in the following.

\subsection{View Generator (Generative Module)}
\begin{figure}
\centering
   \includegraphics[width=1.0\linewidth]{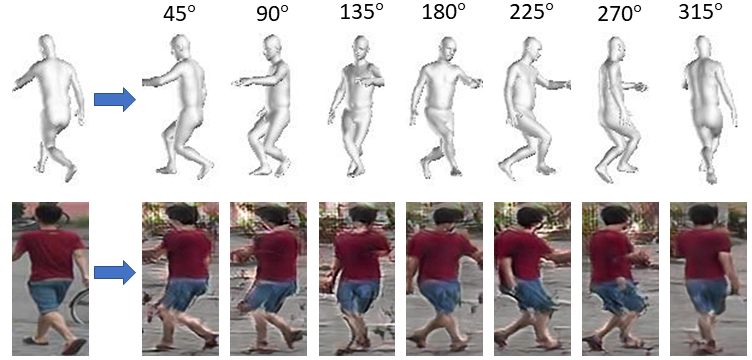}
   \caption{Example images as generated by the View Generator via 3D mesh rotation based on left input image.}
\label{fig:3d meshes rotate}
\end{figure}

As shown in Fig. \ref{fig:general structure}, the proposed View Generator incorporates 4 networks: an identity encoder $E_{id}$, a structure encoder $E_{str}$, a decoder $G$ and an image discriminator $D$. Given an unlabeled person ReID dataset $\mathcal{X}=\{x_{1}, x_{2}, ..., x_{N}\}$, we generate corresponding 3D meshes with a popular 3D mesh generator Human Mesh Recovery (HMR) \cite{kanazawaHMR18}, which simultaneously estimates body shape and pose from a single RGB image. Here, we denote the 2D projection of a 3D mesh as original structure $s_{ori}$. Then, as depicted in Fig. \ref{fig:3d meshes rotate}, we rotate each 3D mesh by $45^{\degree}, 90^{\degree}, 135^{\degree}, 180^{\degree}, 225^{\degree}, 270^{\degree} $ and $315^{\degree}$, respectively and proceed to randomly pick one 2D projection of these rotated meshes as a new structure $s_{new}$. We use the 3D mesh rotation to mimic view-point variance from different cameras. Next, unlabeled images are encoded to identity features by the identity encoder $E_{id}: x \to f_{id}$, while both original and new structures are encoded to structure features by the structure encoder $E_{str}: s_{ori} \to f_{str(ori)}, s_{new} \to f_{str(new)}$. Combining both, identity and structure features, the decoder generates synthesized images $G: (f_{id}, f_{str(ori)}) \to x_{ori}^{\prime}, (f_{id}, f_{str(new)}) \to x_{new}^{\prime}$, where a prime is used to represent generated images. 



Given the lack of real images corresponding to the new structures, we consider a cycle consistency \cite{CycleGAN2017} to reconstruct the original image by swapping the structure features in the View Generator. We encode and decode once again to get synthesized images in original structures  $G(E_{id}(x_{new}^{\prime}), s_{ori}) \to x_{ori}^{\prime\prime}$. We calculate an image reconstruction loss as follows.
\begin{equation}
\begin{split}
\mathcal{L}_{img} = \mathop{\mathbb{E}} [\norm{x-x_{ori}^{\prime}}] + \mathop{\mathbb{E}} [\norm{x-x_{ori}^{\prime\prime}} ]
\end{split}
\label{image_recon_loss}
\end{equation}
In addition, we compute a feature reconstruction loss 
\begin{equation}
\begin{split}
\mathcal{L}_{feat} = &\mathop{\mathbb{E}} [\norm{f_{id}-E_{id}(x_{new}^{\prime})}]+\\
&\mathop{\mathbb{E}} [\norm{f_{id}-E_{id}(x_{ori}^{\prime\prime})} ].
\end{split}
\label{feat_recon_loss}
\end{equation}
The discriminator $D$ attempts to distinguish between real and generated images with the adversarial loss
\begin{equation}
\begin{split}
\mathcal{L}_{adv} = &\mathop{\mathbb{E}} [ \log D(x) + \log (1-D(x_{ori}^{\prime})) ] + \\
&\mathop{\mathbb{E}} [ \log D(x) + \log (1-D(x_{new}^{\prime})) ]  + \\
&\mathop{\mathbb{E}} [ \log D(x) + \log (1-D(x_{ori}^{\prime\prime})) ].
\end{split}
\label{adv_loss}
\end{equation}
Consequently, the overall GAN loss combines the above named losses with weighting coefficients $\lambda_{img}$ and $\lambda_{feat}$
\begin{equation}
\begin{split}
\mathcal{L}_{gan} = \lambda_{img}\mathcal{L}_{img} + \lambda_{feat}\mathcal{L}_{feat}+
   \mathcal{L}_{adv}.
\end{split}
\label{gan_loss}
\end{equation}

\subsection{View Contrast (Contrastive Module)}
The previous reconstruction and adversarial losses work in an unconditional manner. They only explore identity features within the original view-point, which renders appearance representations view-variant. In rotating an original mesh to a different view-point, \eg, from front to side view-point, the generation is prone to fail due to lack of information pertained to the side view. This issue can be alleviated by enhancing the view-invariance of representations. 

Given an anchor image $x$, the first step is to find positive images that belong to the same identity and negative images that belong to different identities. Here, we store all instance representations in a memory bank \cite{Wu2018UnsupervisedFL}, which stabilizes pseudo labels and enlarges the number of negatives during the training with mini-batches. The memory bank $\mathcal{M}$ is updated with a momentum coefficient $\alpha$.
\begin{equation}
\mathcal{M}[i]^t =\alpha\cdot \mathcal{M}[i]^{t-1} + (1-\alpha)\cdot f^t
\label{equ:ema}
\end{equation}
where $\mathcal{M}[i]^t$ and $\mathcal{M}[i]^{t-1}$ respectively refer to the identity feature vector in the $t$ and $t-1$ epochs.

We use a clustering algorithm DBSCAN \cite{Ester1996ADA} on all memory bank feature vectors to generate pseudo identity labels $\mathcal{Y}=\{y_{1}, y_{2}, ..., y_{J}\}$, which are renewed at the beginning of every epoch. 
%
Given the obtained pseudo labels, we have $N_{pos}$ positive and $N_{neg}$ negative instances for each training instance. $N_{pos}$ and $N_{neg}$ vary for different instances. For simplicity in a mini-batch training, we fix common positive and negative numbers for every training instance. Given an image $x$, we randomly sample $K$ instances that have different pseudo identities and one instance representation $f_{pos}$ that has the same pseudo identity with $x$ from the memory bank. Note that $f_{pos}$ is from a random positive image that usually has a pose and camera style different from $x$ and $x_{new}^{\prime}$. $x$ and $x_{new}^{\prime}$ are encoded by $E_{id}$ into identity feature vectors $f$ and $f_{new}^{\prime}$. Next, $f$, $f_{new}^{\prime}$ and $f_{pos}$ are used in turn to form three positive pairs. The $f_{new}^{\prime}$ and $K$ different identity instances in the memory bank are used as $K$ negative pairs. Towards learning robust view-invariant representations, we extend the InfoNCE loss \cite{oord2018representation} into a view-invariant loss between original and generated views. We use $sim(u, v) = \frac{u}{\normsecond{u}} \cdot \frac{v}{\normsecond{v}}$ to denote the cosine similarity. We define the view-invariant loss as a softmax log loss of $K+1$ pairs as following.
\begin{equation}
\begin{split}
  \mathcal{L}_{vi} = \mathop{\mathbb{E}}[\log{(1+\frac{\sum\nolimits_{i=1}^{K}\exp{(sim(f_{new}^{\prime}, k_{i})/\tau)}}{\exp{(sim(f,f_{pos})/\tau)}})}]
\label{view invariant loss}
\end{split}
\end{equation}
\begin{equation}
\begin{split}
  \mathcal{L}_{vi}^{\prime} = \mathop{\mathbb{E}}[\log{(1+\frac{\sum\nolimits_{i=1}^{K}\exp{(sim(f_{new}^{\prime}, k_{i})/\tau)}}{\exp{(sim(f_{new}^{\prime},f)/\tau)}})}]
\label{view invariant loss prime}
\end{split}
\end{equation}
\begin{equation}
\begin{split}
  \mathcal{L}_{vi}^{\prime\prime} = \mathop{\mathbb{E}}[\log{(1+\frac{\sum\nolimits_{i=1}^{K}\exp{(sim(f_{new}^{\prime}, k_{i})/\tau)}}{\exp{(sim(f_{new}^{\prime},f_{pos})/\tau)}})}],
\label{view invariant loss prime prime}
\end{split}
\end{equation}
where $\tau$ indicates a temperature coefficient that controls the scale of calculated similarities. $\mathcal{L}_{vi}$ maximizes the invariance between original and memory positive views. $\mathcal{L}_{vi}^{\prime}$ maximizes the invariance between synthesized and original views. $\mathcal{L}_{vi}^{\prime\prime}$ maximizes the invariance between synthesized and memory positive views. Meanwhile, the synthesized view is pushed away from $K$ negative views in the latent space. Replacing $sim(f_{new}^{\prime}, k_{i})$ in Eq.~\ref{view invariant loss}, Eq.~\ref{view invariant loss prime} and Eq.~\ref{view invariant loss prime prime} with $sim(f, k_{i})$ is another possibility, which pushes away the original view from negative instances. After testing, $sim(f_{new}^{\prime}, k_{i})$ works better, because pushing away the synthesized view from negative instances aid the generation of more accurate synthesized views that look different from the $K$ negative instances.

\subsection{Joint Training}
Our proposed GCL framework is trained in a joint training way. Both GAN and contrastive instance discrimination can be trained in a self-supervised manner. While the GAN learns a data distribution via adversarial learning on each instance, contrastive instance discrimination learns representations by retrieving each instance from candidates. In our designed joint training, the two modules work as two collaborators with the same objective: enhancing the quality of representations built by the shared identity encoder $E_{id}$. We formulate our GCL as an approach to augment contrast for unsupervised ReID. Firstly, the generative module generates online data augmentation, which enhances the positive view diversity for contrastive module. Secondly, the contrastive module, in turn, learns view-invariant representations by matching original and generated views, which refine the generation quality. The joint training boosts both modules simultaneously. Our joint training conducts forward propagation initially on the generative module and subsequently on the contrastive module. Back-propagation is then conducted with an overall loss that combines Eq.~\ref{gan_loss}, Eq.~\ref{view invariant loss}, Eq.~\ref{view invariant loss prime} and Eq.~\ref{view invariant loss prime prime}.
\begin{equation}
   \mathcal{L}_{all} = \mathcal{L}_{gan} + \mathcal{L}_{vi} + \mathcal{L}_{vi}^{\prime} + \mathcal{L}_{vi}^{\prime\prime}
\label{overall loss}
\end{equation}
To accelerate the training process and alleviate the noise from imperfect generation quality at beginning epochs, we need to warm up the four modules used in the View Generator $E_{id}$, $E_{str}$, $G$ and $D$. We firstly use a state-of-the-art unsupervised ReID method to warm up $E_{id}$, which is then considered as a baseline in our ablation studies. Generally speaking, any unsupervised ReID method can be used to warm up $E_{id}$. Before conducting the View Contrast, we freeze $E_{id}$ and warm up $E_{str}$, $G$, and $D$  only with GAN loss in Eq.~\ref{gan_loss} for 40 epochs. In the following, we bring in the memory bank and the pseudo labels to jointly train the whole framework with $\mathcal{L}_{all}$ for another 20 epochs. During the joint training, pseudo labels are updated at the beginning of every epoch. 

\section{Experiments}
\subsection{Datasets and Evaluation Protocols}
Three mainstream person ReID datasets are considered in our experiments, including Market-1501 \cite{Zheng2015ScalablePR}, DukeMTMC-reID \cite{ristani2016MTMC} and MSMT17 \cite{wei2018person}. Market-1501 is composed of 12,936 images of 751 identities for training and 19,732 images of 750 identities for test captured from 6 cameras. DukeMTMC-reID contains 16,522 images of 702 persons for training, 2,228 query images and 17,661 gallery images of 702 persons for test from 8 cameras. MSMT17 is a larger dataset, which contains 32,621 training images of 1,041 identities and 93,820 testing images of 3,060 identities collected from 15 cameras.

Following state-of-the-art unsupervised ReID methods \cite{Wang_2020_CVPR,li2020joint}, we evaluate our proposed method GCL under fully unsupervised setting on the three datasets and under four UDA benchmark protocols, including Market$\to$Duke, Duke$\to$Market, Market$\to$MSMT and Duke$\to$MSMT. We report both quantitative and qualitative results for unsupervised person ReID and view generation.



\begin{table*}
\centering
\scalebox{0.8}{
\begin{tabular}{l|c|c|cccc|c|cccc}
\hline
\multirow{2}{*}{Method} & \multirow{2}{*}{Reference} & \multicolumn{5}{c}{Market1501} & \multicolumn{5}{|c}{DukeMTMC-reID} \\ \cline{3-12}
\multicolumn{1}{c|}{} &\multicolumn{1}{c|}{}&\multicolumn{1}{c|}{Source}& \multicolumn{1}{c}{mAP} & \multicolumn{1}{c}{Rank1} & \multicolumn{1}{c}{Rank5} & \multicolumn{1}{c|}{Rank10} & \multicolumn{1}{c|}{Source}& \multicolumn{1}{c}{mAP} & \multicolumn{1}{c}{Rank1}  & \multicolumn{1}{c}{Rank5} & \multicolumn{1}{c}{Rank10}\\ 
\hline
BUC \cite{Lin2019ABC} &AAAI'19 &None&29.6&61.9&73.5&78.2&None&22.1&40.4&52.5&58.2\\ 
SoftSim \cite{Lin2020UnsupervisedPR} &CVPR'20 &None&37.8&71.7&83.8&87.4&None&28.6&52.5&63.5&68.9\\
TSSL \cite{wu2020tracklet} &AAAI'20&None&43.3&71.2&-&-&None&38.5&62.2&-&-\\
MMCL \cite{Wang_2020_CVPR} &CVPR'20 &None&45.5&80.3&89.4&92.3&None&40.2&65.2&75.9&80.0\\
JVTC \cite{li2020joint}&ECCV'20&None&41.8&72.9&84.2&88.7&None&42.2&67.6&78.0&81.6\\
JVTC+ \cite{li2020joint}&ECCV'20&None&47.5&79.5&89.2&91.9&None&50.7&74.6&82.9&85.3\\
\hline
MMCL* &This paper&None&45.1&79.5&89.0&91.9&None&40.9&64.8&75.2&79.8\\
JVTC* &This paper&None&47.2&75.4&86.7&90.5&None&43.9&66.8&77.6&81.0\\
JVTC+* &This paper&None&50.9&79.1&89.8&92.9&None&52.8&74.9&83.3&85.8\\
ours(MMCL*)&This paper&None&54.9&83.7&91.6&94.0&None&49.3&69.7&79.7&82.8\\
ours(JVTC*) &This paper&None&63.4&83.7&91.6&94.3&None&53.3&72.4&82.0&84.9\\
ours(JVTC+*) &This paper&None&\textbf{66.8}&\textbf{87.3}&\textbf{93.5}&\textbf{95.5}&None&\textbf{62.8}&\textbf{82.9}&\textbf{87.1}&\textbf{88.5}\\
\hline
\hline
ECN \cite{zhong2019invariance}&CVPR'19&Duke&43.0&75.1&87.6&91.6&Market&40.4&63.3&75.8&80.4\\
PDA \cite{li2019cross}&ICCV'19&Duke&47.6&75.2&86.3&90.2&Market&45.1&63.2&77.0&82.5\\
CR-GAN \cite{chen2019instance}&ICCV'19&Duke&54.0&77.7&89.7&92.7&Market&48.6&68.9&80.2&84.7\\
SSG \cite{fu2019self}&ICCV'19&Duke&58.3&80.0&90.0&92.4&Market&53.4&73.0&80.6&83.2\\
MMCL \cite{Wang_2020_CVPR} &CVPR'20 &Duke&60.4&84.4&92.8&95.0&Market&51.4&72.4&82.9&85.0\\
ACT \cite{yang2020asymmetric}&AAAI'20&Duke&60.6&80.5&-&-&Market&54.5&72.4&-&-\\
DG-Net++ \cite{Zou2020JointDA}&ECCV'20&Duke&61.7&82.1&90.2&92.7&Market&63.8&78.9&87.8&90.4\\
JVTC \cite{li2020joint}&ECCV'20&Duke&61.1&83.8&93.0&95.2&Market&56.2&75.0&85.1&88.2\\
ECN+ \cite{zhong2020learning}&PAMI'20&Duke&63.8&84.1&92.8&95.4&Market&54.4&74.0&83.7&87.4\\
JVTC+ \cite{li2020joint}&ECCV'20&Duke&67.2&86.8&95.2&97.1&Market&66.5&80.4&\textbf{89.9}&92.2\\
MMT \cite{ge2020mutual}&ICLR'20&Duke&71.2&87.7&94.9&96.9&Market&65.1&78.0&88.8&\textbf{92.5}\\
CAIL \cite{luo2020generalizing}&ECCV'20&Duke&71.5&88.1&94.4&96.2&Market&65.2&79.5&88.3&91.4\\
\hline
ACT* &This paper&Duke&59.1&78.8&88.9&91.7&Market&51.5&70.9&80.0&83.4\\
JVTC* &This paper&Duke&65.0&85.7&93.6&95.6&Market&56.5&73.9&84.5&87.7\\
JVTC+* &This paper&Duke&67.6&87.0&95.2&97.0&Market&66.7&81.0&\textbf{89.9}&91.5\\
ours(ACT*) &This paper&Duke&66.7&83.9&91.4&93.4&Market&55.4&71.9&81.6&84.6\\
ours(JVTC*) &This paper&Duke&73.4&89.1&95.0&96.6&Market&60.4&77.2&86.2&88.4\\
ours(JVTC+*) &This paper&Duke&\textbf{75.4}&\textbf{90.5}&\textbf{96.2}&\textbf{97.1}&Market&\textbf{67.6}&\textbf{81.9}&88.9&90.6\\
\hline
\end{tabular}}
\caption{Comparison of unsupervised ReID methods (\%) with a ResNet50 backbone on Market and Duke datasets. We test our proposed method on several baselines, whose names are in brackets. * refers to our implementation based on authors' code.} 
\label{table:unsupervised}
\end{table*}

\begin{table}
\centering
\scalebox{0.8}{
\begin{tabular}{l|c|c|cccc}
\hline
\multirow{2}{*}{Method} & \multirow{2}{*}{Reference} & \multicolumn{5}{c}{MSMT17} \\ \cline{3-7}
\multicolumn{1}{c|}{} &\multicolumn{1}{c|}{}&\multicolumn{1}{c|}{Source}& \multicolumn{1}{c}{mAP} & \multicolumn{1}{c}{R1} & \multicolumn{1}{c}{R5} & \multicolumn{1}{c}{R10} \\ 
\hline
MMCL \cite{Wang_2020_CVPR}&CVPR'20&None&11.2&35.4&44.8&49.8\\
JVTC \cite{li2020joint}&ECCV'20&None&15.1&39.0&50.9&56.8\\
JVTC+ \cite{li2020joint}&ECCV'20&None&17.3&43.1&53.8&59.4\\
\hline
JVTC* &This paper&None&13.4&36.0&48.8&54.9\\
JVTC+* &This paper&None&16.3&40.4&55.6&61.8\\
ours(JVTC*) &This paper&None&18.0&41.6&53.2&58.4\\
ours(JVTC+*) &This paper&None&\textbf{21.3}&\textbf{45.7}&\textbf{58.6}&\textbf{64.5}\\
\hline\hline
ECN \cite{zhong2019invariance}&CVPR'19&Market&8.5&25.3&36.3&42.1\\
SSG \cite{fu2019self}&ICCV'19&Market&13.2&31.6&49.6&-\\
MMCL \cite{Wang_2020_CVPR}&CVPR'20&Market&15.1&40.8&51.8&56.7\\
ECN+ \cite{zhong2020learning}&PAMI'20&Market&15.2&40.4&53.1&58.7\\
JVTC \cite{li2020joint}&ECCV'20&Market&19.0&42.1&53.4&58.9\\
DG-Net++ \cite{Zou2020JointDA}&ECCV'20&Market&22.1&48.4&60.9&66.1\\
CAIL \cite{luo2020generalizing}&ECCV'20&Market&20.4&43.7&56.1&61.9\\
MMT \cite{ge2020mutual}&ICLR'20&Market&22.9&49.2&63.1&68.8\\
JVTC+ \cite{li2020joint}&ECCV'20&Market&25.1&48.6&65.3&68.2\\
\hline
JVTC* &This paper&Market&17.1&39.6&53.3&59.3\\
JVTC+* &This paper&Market&20.5&44.0&59.5&71.1\\
ours(JVTC*) &This paper&Market&21.5&45.0&57.1&66.5\\
ours(JVTC+*) &This paper&Market&\textbf{27.0}&\textbf{51.1}&\textbf{63.9}&\textbf{69.9}\\
\hline\hline
ECN \cite{zhong2019invariance}&CVPR'19&Duke&10.2&30.2&41.5&46.8\\
SSG \cite{fu2019self}&ICCV'19&Duke&13.3&32.2&51.2&-\\
MMCL \cite{Wang_2020_CVPR}&CVPR'20&Duke&16.2&43.6&54.3&58.9\\
ECN+ \cite{zhong2020learning}&PAMI'20&Duke&16.0&42.5&55.9&61.5\\
JVTC \cite{li2020joint}&ECCV'20&Duke&20.3&45.4&58.4&64.3\\
DG-Net++ \cite{Zou2020JointDA}&ECCV'20&Duke&22.1&48.8&60.9&65.9\\
MMT \cite{ge2020mutual}&ICLR'20&Duke&23.3&50.1&63.9&69.8\\
CAIL \cite{luo2020generalizing}&ECCV'20&Duke&24.3&51.7&64.0&68.9\\
JVTC+ \cite{li2020joint}&ECCV'20&Duke&27.5&52.9&70.5&75.9\\
\hline
JVTC* &This paper&Duke&19.9&45.4&59.1&64.9\\
JVTC+* &This paper&Duke&23.6&49.4&65.2&71.1\\
ours(JVTC*) &This paper&Duke&24.9&50.8&63.4&68.9\\
ours(JVTC+*) &This paper&Duke&\textbf{29.7}&\textbf{54.4}&\textbf{68.2}&\textbf{74.2}\\
\hline
\end{tabular}}
\caption{Comparison of unsupervised Re-ID methods (\%) with a ResNet50 backbone on MSMT17. * refers to our implementation based on authors' code. }
\label{table:msmt}
\end{table}

\subsection{Implementation Details}
We firstly present network design details of $E_{id}$, $E_{str}$, $G$ and $D$. In the following descriptions, we write the size of feature maps in channel$\times$height$\times$width. Our model design is mainly inspired by \cite{zheng2019joint,Zou2020JointDA}. (1) $E_{id}$ is a ImageNet \cite{Russakovsky2015ImageNetLS} pre-trained ResNet50 \cite{he2016deep} with slight modifications. The original fully connected layer is replaced by a fully connected embedding layer, which outputs identity representations $f$ in 512$\times$1$\times$1 for the View Contrast. In parallel, we add a part average pooling that outputs identity features $f_{id}$ in 2048$\times$4$\times$1 for the View Generator. (2) $E_{str}$ is composed of four convolutional and four residual layers, which output structure features $f_{str}$ in 128$\times$64$\times$32. (3) $G$ contains four residual and four convolutional layers. Every residual layer contains two adaptive instance normalization layers \cite{huang2017arbitrary} that transform $f_{id}$ into scale and bias parameters. (4) $D$ is a multi-scale PatchGAN \cite{isola2017image} discriminator at 64$\times$32, 128$\times$64 and 256$\times$128. 

Then, we present the training and testing configuration details. Our framework is implemented in Pytorch and trained with one Nvidia Titan RTX GPU. (1) For the $E_{id}$ warm-up, we consider JVTC \cite{li2020joint}, because it is a state-of-the-art ReID method that is compatible with both fully unsupervised and UDA settings. We also test other baselines, \eg, MMCL \cite{Wang_2020_CVPR} and ACT \cite{yang2020asymmetric} to demonstrate the generalizability of our method. (2) For training, inputs are resized to 256$\times$128. We empirically set a large weight $\lambda_{img}=\lambda_{feat}=5$ for reconstruction in Eq.~\ref{gan_loss}. With a batch size of 16, we use SGD to train $E_{id}$ and Adam optimizer to train $E_{str}$, $G$ and $D$. Learning rate is set to $1\times 10^{-4}$ during the warm-up. In the joint-training, learning rate in Adam is set to $1\times 10^{-4}$ and $3.5\times10^{-4}$ in SGD and are multiplied by $0.1$ after $10$ epochs. (3) In the View Contrast module, we set the momentum coefficient $\alpha=0.2$ in Eq.~\ref{equ:ema} and the temperature $\tau=0.04$ in Eq.~\ref{view invariant loss}. The number of negatives $K$ is 8192. DBSCAN density radius is set to $2\times10^{-3}$. (4) For testing, only $E_{id}$ is conserved and outputs representations $f$ of dimension 512.

Important parameters are set by a grid search on the fully unsupervised Market-1501 benchmark. The temperature $\tau$ is searched from  $\{0.03,0.04,0.05,0.06,0.07\}$ and finally is set to $0.04$. A smaller $\tau$ increases the scale of similarity scores in the Eq.~\ref{view invariant loss}, Eq.~\ref{view invariant loss prime} and Eq.~\ref{view invariant loss prime prime}, which makes view-invariant losses more sensitive to inter-instance difference. However, when $\tau$ is set to $0.03$, these losses become too sensitive and make the training unstable. The number of negatives $K$ is searched from  $\{2048,4096,8192\}$. A larger $K$ pushes away more negatives in the view-invariant losses. Since the Market-1501 dataset has only 12936 training images, we set $K=8192$. 

\subsection{Unsupervised ReID Evaluation}
\paragraph{Comparison with state-of-the-art methods.}
Tab.~\ref{table:unsupervised} shows the quantitative results on the Market-1501 and DukeMTMC-reID datasets. Tab.~\ref{table:msmt} shows the quantitative results on the MSMT17 dataset. Our method is mainly designed for fully unsupervised ReID. Under this setting, we test the performance of GCL with three different baselines, including MMCL, JVTC and JVTC+. Our implementation of the three baselines provides results that are slightly different from those mentioned in the corresponding papers. Thus, we firstly report results of our implementations and then add our GCL on these baselines. Our method improves the performance of the baselines by large margins. These improvements show that GANs are not limited to cross-domain style transfer for unsupervised ReID. 

Under the UDA setting, we also evaluate the performance of GCL with three different baselines, including ACT, JVTC and JVTC+. The labeled source dataset is only used to warm up our identity encoder $E_{id}$, but not used in our joint generative and contrastive training. Compared to fully unsupervised methods, the UDA warmed $E_{id}$ is stronger and extracts improved identity features. Thus, the performance of UDA methods is generally higher than fully unsupervised methods. With a strong baseline JVTC+, our GCL achieves state-of-the-art performance. 

\paragraph{Ablation Study.}
To better understand the contribution of generative and contrastive modules, we conduct ablation experiments on the two fully unsupervised benchmarks: Market-1501 and DukeMTMC-reID. Quantitative results with a JVTC baseline are reported in Tab.~\ref{table:ablation}. By gradually adding loss functions on the baseline, our ablation experiments correspond to three scenarios. (1) Only Generation: with only $\mathcal{L}_{gan}$, our generation module disentangles identity and structure features. Since there is no inter-view constraint, $E_{id}$ tends to extract view-specific identity features, which decreases the ReID performance. (2) Only Contrast: we use $\mathcal{L}_{vi}^{woGAN}=\mathop{\mathbb{E}}[\log{(1+\frac{\sum\nolimits_{i=1}^{K}\exp{(sim(f,k_{i})/\tau)}}{\exp{(sim(f,f_{pos})/\tau)}}})]$ to train our contrastive module without generation. We also add a set of traditional data augmentation, including random flipping, cropping, jittering, erasing, to train our contrastive module like a traditional memory bank based contrastive method. (3) Joint Generation and Contrast: $\mathcal{L}_{vi}$, $\mathcal{L}_{vi}^{\prime}$ and $\mathcal{L}_{vi}^{\prime\prime}$ enhance the view-invariance of identity representations between original, synthesized and memory-stored positive views, while negative views are pushed away. We provide how view-invariant representations learned from generated views affect pseudo labels in Appendix~\ref{Appendix: Effects on pseudo labels}.

\begin{figure}
\centering
   \includegraphics[width=0.9\linewidth]{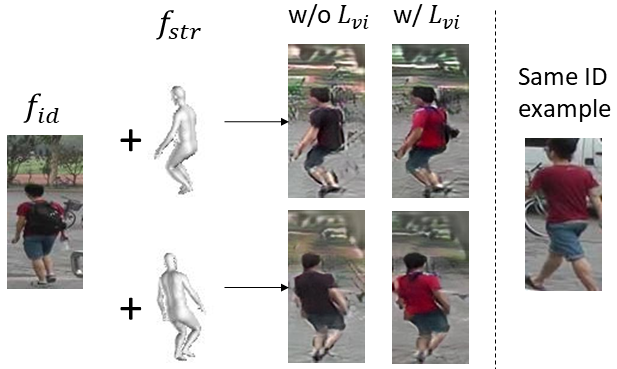}
   \caption{Qualitative ablation study on the view-invariant losses. For simplicity, $\mathcal{L}_{vi}$ denotes three view-invariant losses $\mathcal{L}_{vi}+\mathcal{L}_{vi}^{\prime}+\mathcal{L}_{vi}^{\prime\prime}$, which helps $E_{id}$ to extract view-invariant features (red shirt). }
\label{fig:qualitative view-invariant}
\end{figure}

We also conduct a qualitative ablation study, where synthesized novel views without and with view-invariant losses are illustrated in Fig. \ref{fig:qualitative view-invariant}. Results confirm that $E_{id}$ extracts view-specific identity features (black bag), in the case that view-invariant losses are not used. Given view-invariant losses, $E_{id}$ is able to extract view-invariant identity features (red shirt). Another example is provided in Appendix~\ref{Appendix: View-invariant losses}.

\begin{table}
\centering
\scalebox{0.8}{
\begin{tabular}{c|cc|cc}
\hline
\multirow{2}{*}{Loss}  & \multicolumn{2}{c}{Market-1501} & \multicolumn{2}{|c}{DukeMTMC-reID} \\ \cline{2-5}
\multicolumn{1}{c|}{} & \multicolumn{1}{c}{mAP} & \multicolumn{1}{c|}{Rank1} & \multicolumn{1}{c}{mAP} & \multicolumn{1}{c}{Rank1} \\ \hline
Baseline&47.2&75.4&43.9&66.8\\
$+\mathcal{L}_{gan}$&41.6&69.0&25.8&45.9\\
$+\mathcal{L}_{vi}^{woGAN}$&47.8&75.2&44.1&67.8\\
$+\mathcal{L}_{vi}^{woGAN}+TDA$&53.7&78.7&48.5&70.0\\
$+\mathcal{L}_{gan}+\mathcal{L}_{vi}$&54.1&79.4&47.4&68.4\\
$+\mathcal{L}_{gan}+\mathcal{L}_{vi}+\mathcal{L}_{vi}^{\prime}$&59.2&82.2&50.5&71.0\\
$+\mathcal{L}_{gan}+\mathcal{L}_{vi}+\mathcal{L}_{vi}^{\prime}+\mathcal{L}_{vi}^{\prime\prime}$&\textbf{63.4}&\textbf{83.7}&\textbf{53.3}&\textbf{72.4}\\
\hline
\end{tabular}}
\caption{Ablation study on loss functions used in two modules. (1). $\mathcal{L}_{gan}$ corresponds to generation w/o contrast. (2). $\mathcal{L}_{vi}^{woGAN}$ corresponds to contrast w/o generation. TDA denotes traditional data augmentation. (3). $\mathcal{L}_{gan}+\mathcal{L}_{vi}$ ($\mathcal{L}_{vi}^{\prime}$ and $\mathcal{L}_{vi}^{\prime\prime}$) correspond to joint generative and contrastive learning. }
\label{table:ablation}
\end{table}

\subsection{Generation Quality Evaluation}
\paragraph{Comparison with state-of-the-art methods.}
\begin{figure*}
\centering
   \includegraphics[width=1.0\linewidth]{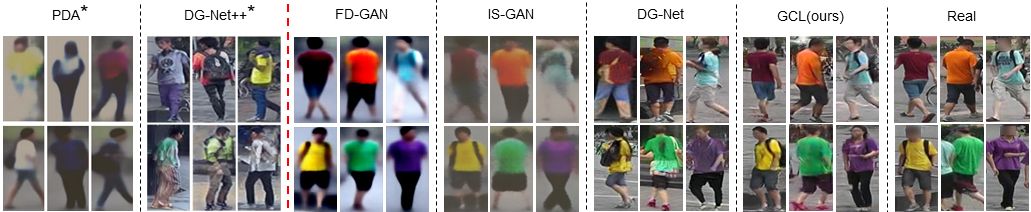}
\caption{Comparison of the generated images on Market-1501 dataset. $\star$ refers to methods without sharing source code, whose examples are cropped from their papers. Examples of FD-GAN, IS-GAN, DG-Net and GCL are generated from six real images shown in the figure.}
\label{fig:qualitative comparison}
\vspace{-10pt}
\end{figure*}

\begin{table}
\centering
\scalebox{0.85}{
\begin{tabular}
{ccc}
\hline
Method& FID(realism) & SSIM(diversity) \\
\hline
Real & \textbf{7.22}& 0.350 \\\hline
FD-GAN \cite{ge2018fd} & 216.88 & 0.271 \\
IS-GAN \cite{NIPS2019_8771} & 281.63 & 0.165 \\
DG-Net \cite{zheng2019joint}& 18.24 & 0.360 \\\hline
Ours(U) & 59.86& 0.367\\
Ours(UDA) & 53.07& \textbf{0.369} \\
\hline
\end{tabular}}
\caption{Comparison of FID (lower is better) and SSIM (higher is better) on Market-1501 dataset. U denotes the fully unsupervised setting. UDA denotes Duke$\to$Market setting. }
\label{tab:FID}
\end{table}

We compare generated images between our proposed GCL under the JVTC \cite{li2020joint} warmed fully unsupervised setting and state-of-the-art GAN-based ReID methods in Fig. \ref{fig:qualitative comparison}. FD-GAN \cite{ge2018fd}, IS-GAN \cite{NIPS2019_8771} and DG-Net \cite{zheng2019joint} are supervised Re-ID methods. Since the source code of these three methods is available, we compare generated images of same identities. We observe that there exists blur in images generated by FD-GAN and IS-GAN. DG-Net generates sharper images, but different body shapes and some incoherent objects (bags and clothes) are observed. PDA \cite{li2019cross} and DG-Net++ \cite{Zou2020JointDA} are UDA methods, whose source code is not yet released. We can only compare several generated images with unknown identities as illustrated in their papers. PDA generates blurred cross-domain images, whose quality is similar to FD-GAN and IS-GAN. DG-Net++ extends DG-Net into cross-domain generation, which has same problems of body shape and incoherent objects. Our GCL preserves better body shape information and does not generate incoherent objects. Moreover, our GCL is a fully unsupervised method.

We use Fr\'echet Inception Distance (FID) \cite{heusel2017gans} to measure visual quality, as well as Structural SIMilarity (SSIM) \cite{wang2004image} to capture structure diversity of generated images. In Tab.~\ref{tab:FID}, we compare our method with FD-GAN \cite{ge2018fd}, IS-GAN \cite{NIPS2019_8771} and DG-Net \cite{zheng2019joint}, whose source code is available. FID measures the distribution distance between generated and real images, where a lower FID represents the case, where generated images are similar to real ones. SSIM measures the intra-class structural similarity, where a larger SSIM represents a larger diversity. We note that DG-Net is outperforms our method w.r.t. FID, because the distribution is better maintained with ground truth identities in the supervised method DG-Net. However, our method is superior to DG-Net w.r.t. SSIM, as DG-Net swaps intra-dataset structures, whereas our rotated meshes build structures that do not exist in the original dataset. 

\begin{figure}
\centering
   \includegraphics[width=1\linewidth]{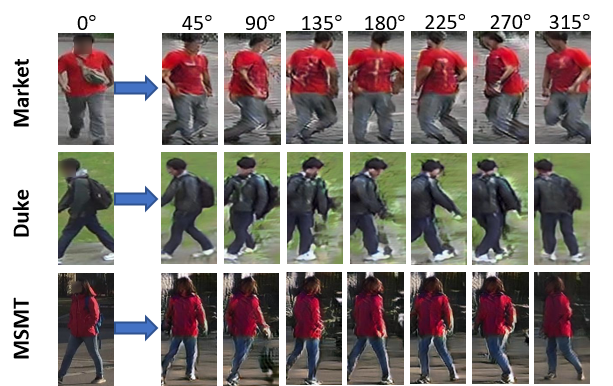}
   \caption{Generated novel views on the three datasets.}
\label{fig:qualitative duke msmt}
\vspace{-10pt}
\end{figure}
 
\begin{figure}
\centering
   \includegraphics[width=1.0\linewidth]{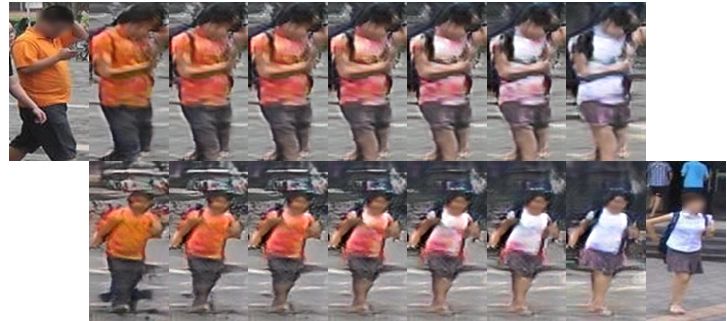}
   \caption{Linear interpolation on identity features. Identity features are swapped between left and right persons. }
\label{fig:interpolate}
\end{figure}

\paragraph{More discussion.} 
To validate, whether identity and structure features can be really disentangled under a fully unsupervised ReID setting, two experiments are conducted by changing firstly only structure features and then only identity features. Results in Fig. \ref{fig:qualitative duke msmt} show that changing structure features only change structures and do not affect appearances. We also fix structure features and linearly interpolate two random identity feature vectors. Results in Fig. \ref{fig:interpolate} show that identity features only change appearances and do not affect structures in generated images. More examples are provided in Appendix~\ref{Appendix: Generated views}.

\section{Conclusions}
In this paper, we propose a joint generative and contrastive learning framework for unsupervised person ReID. Deviating from previous contrastive methods with traditional data augmentation techniques, we generate diversified views with a 3D mesh guided GAN. These generated novel views are then combined with original images in memory based contrastive learning, in order to learn view-invariant representations, which in turn improve generation quality. Our generative and contrastive modules mutually promote each other's performance in unsupervised ReID. Moreover, our framework does not rely on a source dataset, which is mandatory in style transfer based methods. Extensive experiments on three datasets validate the effectiveness of our framework in both unsupervised person ReID and multi-view person image generation. 



{\small
\bibliographystyle{ieee_fullname}
\balance
\bibliography{egbib}
}


\begin{appendices}
\nobalance

\section{Cycle consistency} \label{Appendix: Cycle consistency}
Two kinds of cycle consistency are used in our proposed method GCL. 1) Based on a popular assumption in ReID that two images of a same identity should have same nearest neighbors in the dataset, we have calculated $k$-reciprocal Jaccard distance \cite{zhong2017re} for the DBSCAN clustering. This operation effectively makes pseudo labels more reliable for contrastive learning. 2) Since no paired data are available, we have used the CycleGAN \cite{CycleGAN2017} structure to supervise the generative module. By minimizing the image and feature reconstruction losses in a cycle consistency, representations are disentangled into appearance and structure features, which permits generating person images in novel view-points without changing identity information. 

\section{View-invariant losses} \label{Appendix: View-invariant losses}
\begin{figure}[t]
\centering
  \includegraphics[width=1\linewidth]{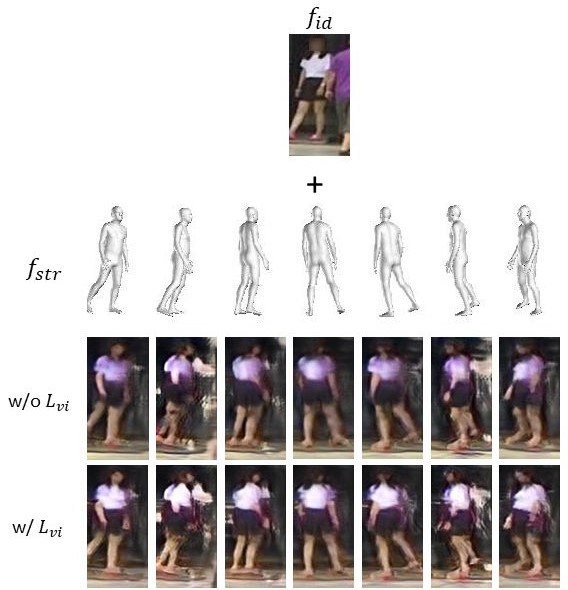}
  \caption{More qualitative ablation study on the view-invariant losses. For simplicity, $\mathcal{L}_{vi}$ denotes three view-invariant losses $\mathcal{L}_{vi}+\mathcal{L}_{vi}^{\prime}+\mathcal{L}_{vi}^{\prime\prime}$, which helps $E_{id}$ to extract better identity features (white shirt).}
\label{fig:more qualitative ablation study}
\end{figure}

We illustrate another example in Fig.~\ref{fig:more qualitative ablation study} to confirm the effectiveness of the view-invariant losses in generation. When GCL is trained without the view-invariant losses, GCL degrades to a traditional CycleGAN, which is prone to affected by the noise inside the original image. The view-invariant losses help the identity encoder to extract identity features shared between different views, which are robust to the noise inside the original image. 

\section{Effects on pseudo labels} \label{Appendix: Effects on pseudo labels}
\begin{figure}[t]
\centering
  \includegraphics[width=0.9\linewidth]{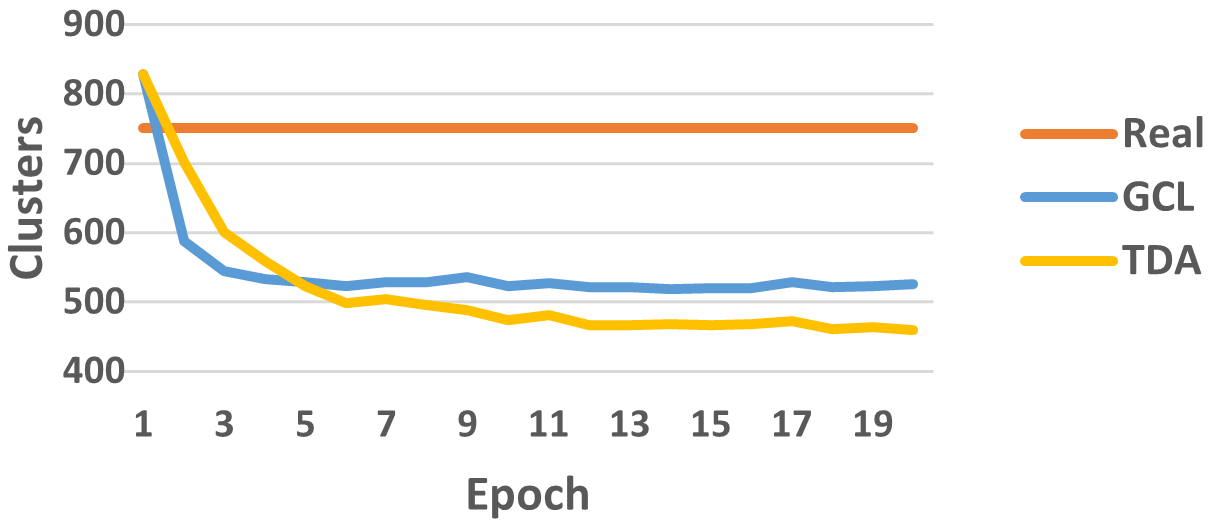}
  \caption{Cluster number curve on Market-1501. TDA denotes traditional data augmentation, including random flipping, cropping, jittering, erasing.}
\label{fig:cluster curve}
\end{figure}

We minimize intra-class variance via contrasting generated images, which leads to a larger inter-class distance in latent space. Learning view-invariant representations from diversified generated data helps clustering algorithms to generate more accurate pseudo labels. With a same DBSCAN clustering, the cluster number of GCL is closer to real identity number than that of contrastive learning with traditional data augmentation. For example, Market-1501 dataset has 751 real identities. DBSCAN in GCL categorizes unlabeled images into around 520 clusters, while the contrastive learning with traditional data augmentation has around 460 clusters (see Fig. \ref{fig:cluster curve}).

\section{Generated views} \label{Appendix: Generated views}
\begin{figure}
\centering
   \includegraphics[width=1\linewidth]{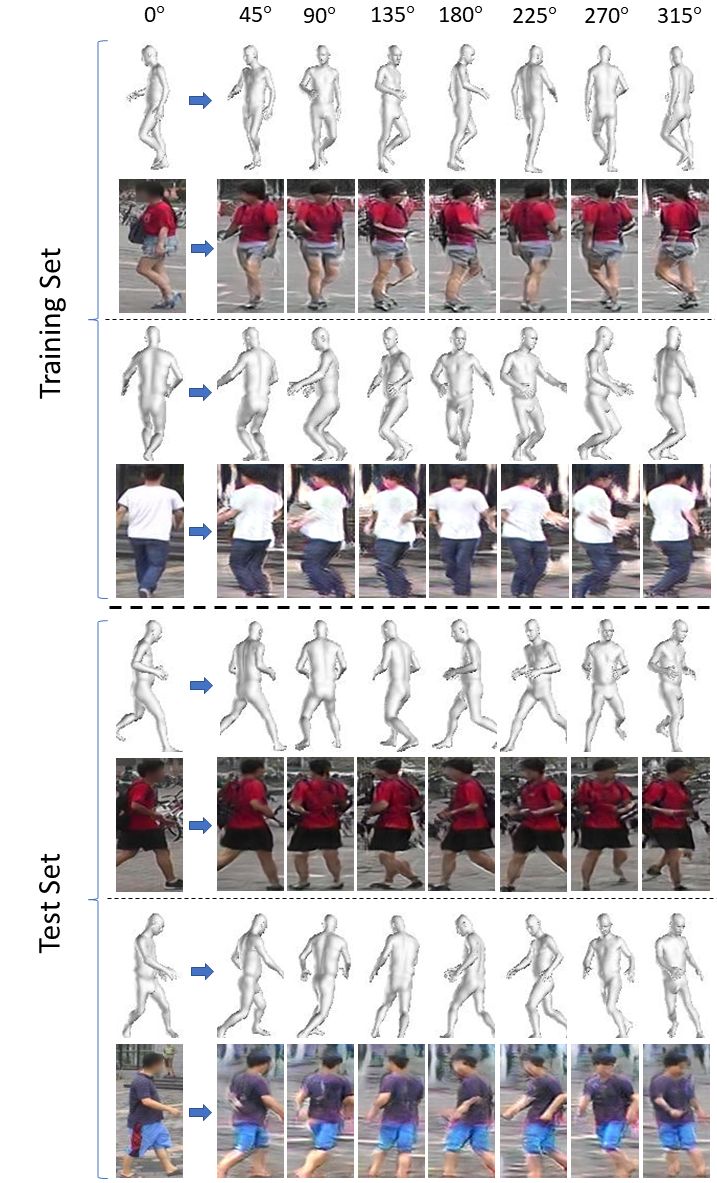}
   \caption{Examples of generated novel views on \textbf{Market-1501} training and test sets.}
\label{fig:market}
\end{figure}

\begin{figure}
\centering
   \includegraphics[width=1\linewidth]{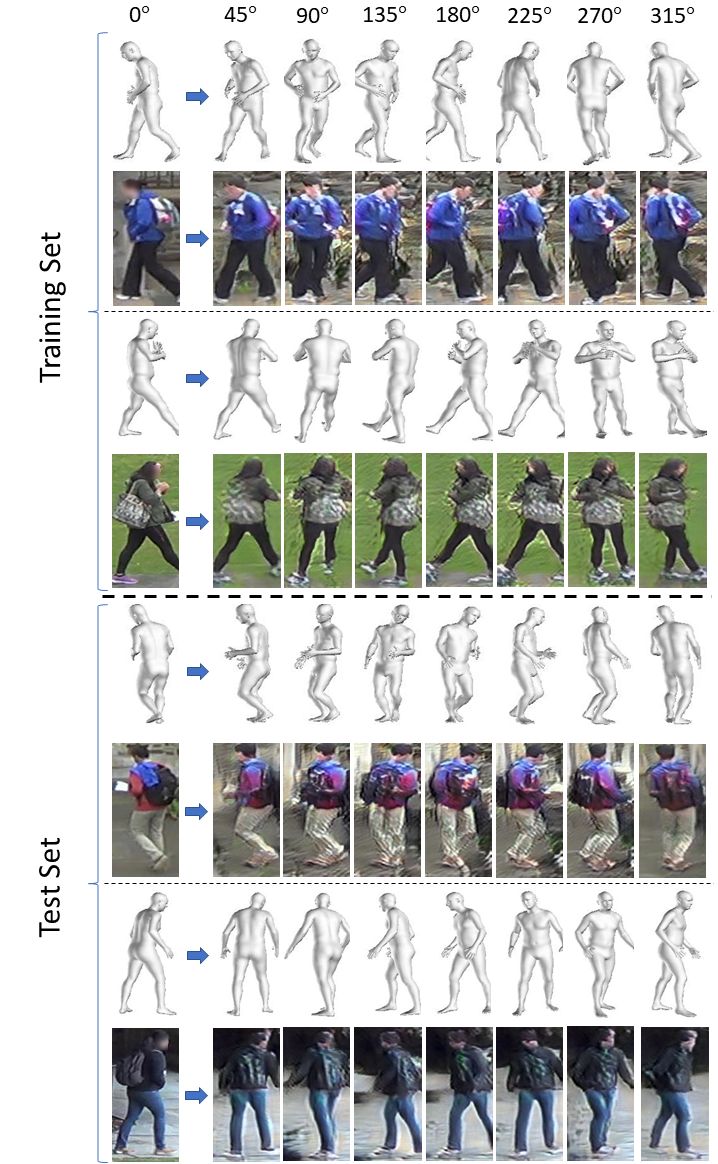}
   \caption{Examples of generated novel views on \textbf{DukeMTMC-reID} training and test sets.}
\label{fig:duke}
\end{figure}

\begin{figure}
\centering
   \includegraphics[width=1\linewidth]{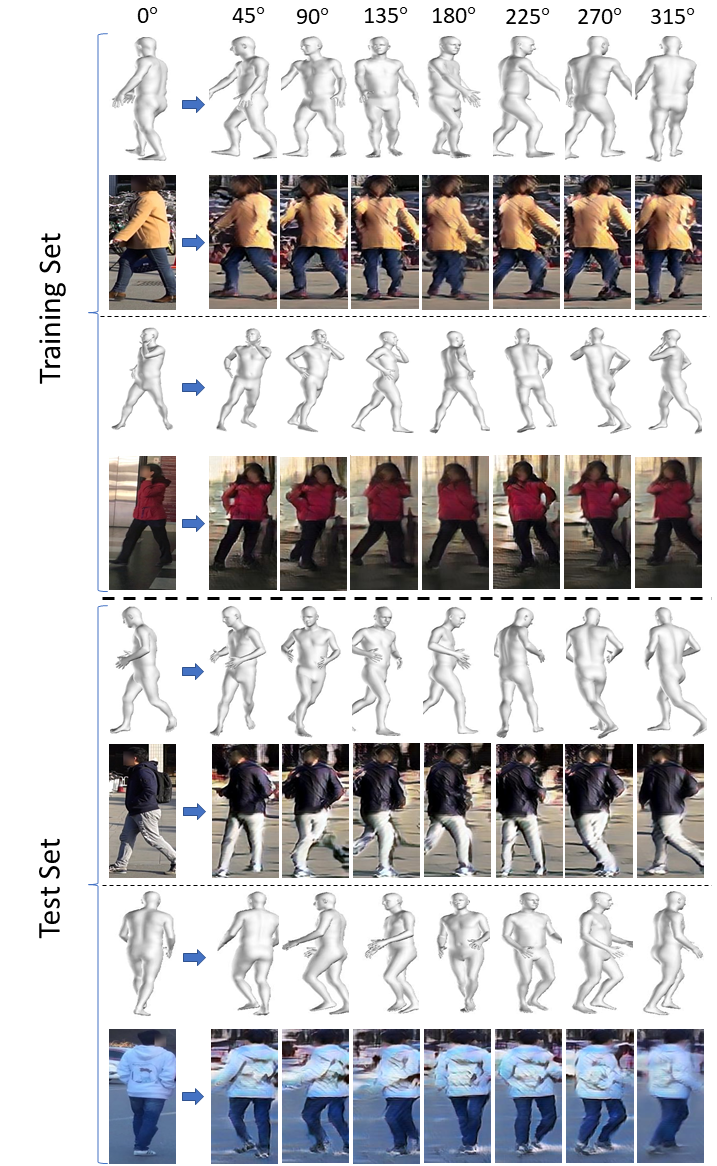}
   \caption{Examples of generated novel views on \textbf{MSMT17} training and test sets.}
\label{fig:msmt}
\end{figure}

We illustrate more examples of generated views with a JVTC \cite{li2020joint} fully unuspervised baseline on Market-1501 in Fig.~\ref{fig:market}, DukeMTCM-reID in Fig.~\ref{fig:duke} and MSMT17 in Fig.~\ref{fig:msmt}. Here, we show generated examples from both training and test sets to confirm the effectiveness of our GCL. Generally, the generation quality is good enough to help our GCL learn view-invariant representations. However, there are still some limitations, \eg, some visual blurs still exist and detailed identity information is lost in some cases (in the \textit{bottom row} of Fig.~\ref{fig:market}, the red logo on the shorts disappears in the generated images). In future work, we believe that the visual blurs can be alleviated by leveraging the architectures from more recent GANs~\cite{brock2018large, Karras2019stylegan2} in our generator and detailed identity information can be better preserved when better unsupervised baselines are available. 

Generally, it is easier to generate novel views of $45\degree$, $180\degree$ and $315\degree$. $45\degree$ and $315\degree$ are small rotations, in which original and synthesized images can share maximal identity information. $180\degree$ can be roughly regarded as a horizontal flipping. Results in Tab.~\ref{tab:view FID} verify this supposition. Our generated novel views on the three datasets will be released as a new dataset to facilitate future research on view-invariant and unsupervised ReID.

\begin{table}
\centering
\scalebox{0.8}{
\begin{tabular}
{ccccccccc}
\hline
 & $45^{\degree}$ & $90^{\degree}$ & $135^{\degree}$ & $180^{\degree}$ & $225^{\degree}$ & $270^{\degree}$ & $315^{\degree}$ \\
\hline
Market  & 56.55 & 75.83 & 62.59 & 51.22 & 62.31 & 70.79 & 55.00  \\
Duke &58.24 &72.21 & 66.29& 57.41&64.61 & 68.08& 55.03\\
MSMT &54.14 &64.46 &60.75& 55.98&59.78 &62.26 &48.40 \\
\hline
\end{tabular}}
\caption{FID score on different views.}
\label{tab:view FID}
\end{table}

\end{appendices}

\end{document}